\documentclass[sigconf]{acmart}

\usepackage{hyperref}
\usepackage{subcaption}
\usepackage{enumitem}
\usepackage{makecell}
\usepackage[nameinlink]{cleveref}
\usepackage{multirow}
\usepackage{threeparttable}
\usepackage{xcolor}
\usepackage{siunitx}
\usepackage{fancyvrb}
\usepackage{pifont}
\usepackage{xspace}
\usepackage{balance}
\usepackage{adjustbox}
\usepackage{xurl}
\usepackage{colortbl}
\usepackage{amsmath}
\usepackage{tcolorbox}

\definecolor{framecolor}{RGB}{0,0,0}
\definecolor{backcolor}{RGB}{245,245,245}
\definecolor{titlecolor}{RGB}{70,70,70}

\newcommand{\wcircle}[1]{\ding{\numexpr171 + #1}}
\newcommand{\bcircle}[1]{\ding{\numexpr181 + #1}}

\usepackage{mdframed}
\newmdenv[skipabove=10pt,skipbelow=10pt,leftmargin=0pt,rightmargin=0pt,linewidth=1pt,linecolor=black,backgroundcolor=gray!10]{formal}
\usepackage{framed}
\usepackage{changepage}
\usepackage{newfloat}
\usepackage{listings}

\begin{document}

\title{Using Contrastive Learning to Improve Two-Way Reasoning in Large Language Models: The Obfuscation Task as a Case Study}

\author{Serge Lionel NIKIEMA}
\orcid{0009-0001-0066-3694}
\affiliation{%
  \institution{University of Luxembourg}
  \city{Luxembourg}
  \country{Luxembourg}
}
\email{lionel.nikiema@uni.lu}

\author{Jordan Samhi}
\orcid{0000-0001-6052-6184}
\affiliation{%
  \institution{University of Luxembourg}
  \city{Luxembourg}
  \country{Luxembourg}
}
\email{jordan.samhi@uni.lu}

\author{Micheline Bénédicte Moumoula}
\orcid{}
\affiliation{%
  \institution{University of Luxembourg}
  \city{Luxembourg}
  \country{Luxembourg}
}
\email{micheline.moumoula@uni.lu}

\author{Albérick Euraste Djiré}
\orcid{}
\affiliation{%
  \institution{University of Luxembourg}
  \city{Luxembourg}
  \country{Luxembourg}
}
\email{euraste.djire@uni.lu}

\author{Abdoul Kader Kaboré}
\orcid{0000-0002-3151-9433}
\affiliation{%
  \institution{University of Luxembourg}
  \city{Luxembourg}
  \country{Luxembourg}
}
\email{abdoulkader.kabore@uni.lu}

\author{Jacques Klein}
\orcid{0000-0003-4052-475X}
\affiliation{%
  \institution{University of Luxembourg}
  \city{Luxembourg}
  \country{Luxembourg}
}
\email{jacques.klein@uni.lu}

\author{Tegawendé F. Bissyandé}
\orcid{0000-0001-7270-9869}
\affiliation{%
  \institution{University of Luxembourg}
  \city{Luxembourg}
  \country{Luxembourg}
}
\email{tegawende.bissyande@uni.lu}

\begin{abstract}

This research addresses a fundamental question in AI: whether large language models truly understand concepts or simply recognize patterns. The authors propose "bidirectional reasoning"—the ability to apply transformations in both directions without being explicitly trained on the reverse direction—as a test for genuine understanding. They argue that true comprehension should naturally allow reversibility, such as a model that can change a variable name like "userIndex" to "i" should also be able to infer that "i" represents a user index without reverse training. 
The researchers tested current language models and discovered what they term "cognitive specialization": when models are fine-tuned on forward tasks, their performance on those tasks improves but their ability to reason bidirectionally becomes significantly worse. 

To address this issue, they developed Contrastive Fine-Tuning (CFT), which trains models using three types of examples: positive examples that maintain semantic meaning, negative examples with different semantics, and forward-direction obfuscation examples. This approach aims to develop deeper understanding rather than surface-level pattern recognition and allows reverse capabilities to develop naturally without explicit reverse training. 
Their experiments demonstrated that CFT successfully achieved bidirectional reasoning, enabling strong reverse performance while maintaining forward task capabilities. The authors conclude that bidirectional reasoning serves both as a theoretical framework for assessing genuine understanding and as a practical training approach for developing more capable AI systems. 
\end{abstract}

\settopmatter{printacmref=false}
\setcopyright{none}
\renewcommand\footnotetextcopyrightpermission[1]{}

\maketitle

\section{Introduction}
\label{sec:introduction}

The rapid advancement of Large Language Models (LLMs) has revolutionized automated code understanding and generation.
Models now demonstrate impressive capabilities in tasks ranging from code completion~\cite{chen2021evaluating, husein2024large} to bug fixing~\cite{zubair2024use, bouzenia2024repairagent} and cross-language translation~\cite{eniser2025towards, pan2024lost}. 
However, a fundamental question persists: do these  models truly understand code (i.e., its semantic),
or do they merely replicate patterns observed during training?

This distinction is critical for evaluating the \emph{reliability} and \emph{generalizability} of LLMs in software engineering applications, particularly as these systems are increasingly deployed in high-stakes real-world environments where genuine semantic understanding is paramount.

Recent work has revealed concerning limitations in LLM code comprehension. 
~\cite{nikiema2025code} demonstrated that recent state-of-the-art models struggle with basic code understanding tasks when evaluated beyond surface-level pattern matching. 
Studies on adversarial robustness have shown that code models can be easily fooled by simple transformations that preserve semantic meaning but alter syntactic structure~\cite{zhou2019devign, zhang2020generating}. 
These findings suggest that current models may be more brittle than their benchmark performance  indicate.

To investigate this question systematically, we propose the \textbf{Bidirectional Reasoning Hypothesis (BRH)} as a formal test for code semantic understanding.
To that end, we consider obfuscation as a controlled transformation since it maintains code semantic after transformation.
Our central premise is that true comprehension of a code semantic should enable bidirectional reasoning: if a model genuinely understands the principles underlying a transformation $T$ (such as code obfuscation), it should be capable of performing the inverse transformation $T^{-1}$ (deobfuscation) without explicit training on reverse examples, \emph{provided the transformation preserves program semantics}.
This reversibility test provides a framework for distinguishing between sophisticated pattern matching and genuine semantic understanding.

Obfuscation techniques deliberately modify code structure but preserve functional semantics~\cite{collberg2009surreptitious, schrittwieser2016protecting}, making them ideal for testing whether models grasp underlying program logic or merely surface-level patterns. 
Through systematic evaluation across six state-of-the-art models and three obfuscation techniques, we document a phenomenon we term \textbf{cognitive specialization}: when fine-tuning models on forward transformations (i.e., the obfuscation task), it not only fails to enable reverse capabilities (i.e., deobfuscating code) but also actively degrades pre-existing bidirectional reasoning present in base models.

To address these limitations, we utilize the
\textbf{Contrastive Fine-Tuning (CFT)} technique (already used in vision learning \cite{chen2020simple}, and are the first to apply this technique on code comprehension.
To this end, we fine-tuned models with triplets consisting of:  
\wcircle{1} non-obfuscated--obfuscated code pairs with the same semantics,  
\wcircle{2} non-obfuscated--non-obfuscated code pairs with different semantics, and  
\wcircle{3} obfuscated code examples for forward transformation learning.
This contrastive approach forces the development of principled understanding rather than superficial pattern matching, enabling reverse capabilities to emerge \emph{naturally} without explicit reverse training.

Our contributions are threefold: 
\bcircle{1} We introduce the Bidirectional Reasoning Hypothesis as a formal framework for evaluating genuine understanding in code transformation tasks. 
\bcircle{2} We identify and characterize cognitive specialization as a fundamental limitation of standard fine-tuning approaches. 
\bcircle{3} We propose and validate Contrastive Fine-Tuning as a practical solution that enables robust bidirectional reasoning while maintaining forward task performance. Our experimental results demonstrate a breakthrough in reversibility capabilities: while standard fine-tuning achieves 0\% success on deobfuscation tasks, CFT enables 39-52\% reverse performance across multiple models and transformation types.

\section{Bidirectional Reasoning Framework}
\label{sec:theorical}

In this section, we introduce the Bidirectional Reasoning Framework, which provides a principled basis for distinguishing genuine semantic understanding from mere pattern matching in LLMs.

\subsection{Semantic Understanding vs. Pattern Matching}

A fundamental challenge in evaluating LLMs is distinguishing between genuine semantic understanding and pattern replication. 
We postulate that \textbf{true understanding of code transformations requires bidirectional reasoning}, i.e., the ability to perform both forward and inverse operations, \emph{unless the transformation is inherently irreversible}.

Consider this intuition: if one truly understands multiplication, one understands division since the two are inherently linked.
Similarly, comprehension of code obfuscation should enable deobfuscation without additional training, only if the transformation preserves program semantics.

\subsection{Formal Framework}

\textbf{Definition 1 (Bidirectional Reasoning):} For a given transformation $T$, genuine understanding requires proficiency in both directions, whenever $T$ is semantically reversible:  
\begin{itemize}
    \item \textbf{Forward:} $T: \text{Code}_{\text{original}} \rightarrow \text{Code}_{\text{transformed}}$
    \item \textbf{Reverse:} $T^{-1}: \text{Code}_{\text{transformed}} \rightarrow \text{Code}_{\text{original}\sim}$
    \item \textbf{Consistency:} The semantic meaning remains preserved across both operations
\end{itemize}
where $\text{Code}_{\text{original}\sim}$ represents a version of $\text{Code}_{\text{original}}$ that is semantically equivalent.

\noindent
\textbf{Definition 2 (Cognitive Specialization):} A learning pathology where training on $T$ creates directional bias, degrading performance on $T^{-1}$. This occurs when models optimize for surface-level pattern replication rather than semantic understanding~\cite{kirkpatrick2017overcoming,french1999catastrophic}.

\noindent
\textbf{Definition 3 (Understanding Criterion):} 
We measure genuine understanding through bidirectional capability—the simple principle that truly understanding a transformation means you can do it in both directions.

If a model can obfuscate code but cannot deobfuscate it, this reveals pattern matching rather than understanding. Genuine comprehension requires both forward performance (F) and reverse performance (R) to be strong. We evaluate understanding by testing whether models trained only on obfuscation can also perform deobfuscation without explicit reverse training.

This bidirectional test provides a clear, objective criterion: models that achieve high performance in both directions demonstrate genuine understanding, while those that excel only in the trained direction reveal sophisticated memorization.
\subsection{Central Hypothesis}

\textbf{Current approaches suffer from cognitive specialization:} Traditional fine-tuning creates unidirectional neural pathways that excel at forward transformation but fail severely at reverse reasoning~\cite{mccloskey1989catastrophic,goodfellow2013empirical}. 
This phenomenon mirrors catastrophic forgetting~\cite{kirkpatrick2017overcoming}, where task-specific optimization eliminates previously acquired capabilities.

\textbf{Our solution: Contrastive Fine-Tuning.}  
CFT mitigates cognitive specialization by training on transformation triplets—positive (semantics preserved), negative (semantics altered), and forward obfuscation examples.  
This contrastive setup compels models to build bidirectional semantic representations rather than memorizing surface patterns, thereby enabling reverse capabilities to emerge naturally~\cite{chen2020simple,hadsell2006dimensionality}.

\section{Experimental Methodology}
\label{sec:experiment_setup}

\subsection{Large Language Models Considered}

Our study evaluates both open-source and API-based LLMs. Hereafter we describe which LLMs were considered:

\begin{itemize}
    \item \textbf{Open-Source Models:} We include models in the 7B–15B parameter range, which represent a practical balance between computational feasibility and state-of-the-art performance in code-specific tasks. More specifically, we considered Qwen2.5-7B, Qwen2.5-Coder-7B, DeepSeek-R1.5-7B, Mistral-7B, and StarCoder-15B.

    \item \textbf{API-Based Models:} We evaluate GPT-3.5-Turbo and GPT-4.1-mini-2025-04-14, widely used proprietary models that provide competitive baselines and enable comparison between open and closed ecosystems.
\end{itemize}

\subsection{Dataset Collection and Processing}

\subsubsection{Base Code Repository.}
To build our dataset, we collected \num{10000} Java programs from CodeNet~\cite{puri2021codenet}.
with the following criteria: 
\wcircle{1} we randomly  sampled across different problem domains,
\wcircle{2} we only kept programs that successfully compiled and run, and 
\wcircle{3} we only kept programs for which a test suite was available. 

\noindent
\textbf{Why 10\,000 samples?}
We employ \num{10000} samples for open-source models based on convergence studies demonstrating that 7B--16B parameter models reach a performance plateau after \num{10000}--\num{20000} examples, with optimal improvements occurring within the \num{5000}--\num{15000} range for code-specific tasks~\cite{chen2021evaluating,nijkamp2023codegen}. 
For API-based models (GPT-3.5-Turbo, GPT-4.1-Mini), we utilize \num{1000} high-quality pairs following Microsoft's empirical validation that ``complex tasks require \num{1000}'s of high quality training examples''~\cite{dettmers2023qlora} and OpenAI's demonstrated 4$\times$ quality improvement when scaling from 50 to \num{1000} samples~\cite{hu2022lora}.


\subsubsection{Transformation Generation.}

To evaluate bidirectional reasoning, we require controlled transformations that preserve program semantics while altering surface-level structure.  
Such transformations create a reliable testbed to distinguish genuine semantic understanding from mere pattern replication.  
We employ the Obfuscator \cite{superblaubeere27obfuscator} obfuscation tool to generate three distinct obfuscated versions of each program:  
\begin{itemize}
    \item \textbf{Variable Renaming:} Systematic obfuscation of identifiers without altering functionality.  
    \item \textbf{Dead Code Insertion:} Injection of non-functional statements to change syntax while preserving semantics.  
    \item \textbf{String Encryption:} Encryption of string literals with embedded decryption logic, which maintains runtime behavior.
\end{itemize}
These transformations ensure that the meaning (i.e., the semantic) of the program remains unchanged, which allows to rigorously test whether models grasp underlying semantics or simply rely on surface patterns.
At the end of this process, we obtain three dedicated datasets, one for each transformation type, each sample is a pair with the original code and the obfuscated version.  

\subsubsection{Final Dataset Composition.}
To ensure a balanced and reliable evaluation, we constructed the dataset as follows.  
For open-source models, we used \num{10000} (original--obfuscated) program pairs per transformation type.  
For API-based models, we employed \num{1000} (original--obfuscated) program pairs per transformation type also.
In addition, we included an \textbf{Evaluation Set} of \num{300} curated samples from the  \cite{li2023condefects} dataset with verified test coverage.  

We selected the CodeNet and ConDefects datasets because they provide compilable code, which is essential for applying the obfuscation process and ensuring reliable test case coverage.



\subsection{Evaluation Metrics.}
We employ semantic correctness metrics (compilation \\ rate, execution success, test case pass rate), syntactic similarity via CodeBLEU~\cite{ren2020codebleu}, bidirectional consistency measures, and transfer capability assessments. 
CodeBLEU serves as our primary syntactic metric due to its superior correlation with human evaluation ~\cite{ren2020codebleu}. 

For code readability evaluation, we utilize the readability assessment framework~\cite{scalabrino2018readability} that provides standardized readability scores (0-1 scale) based on code complexity, variable naming clarity, and structural organization. 

\subsection{Fine-Tuning Methodologies}
We employ LoRA~\cite{hu2022lora} for open-source models, which updates only small additional matrices instead of modifying the entire model. This reduces trainable parameters by \num{10000}$\times$ while preventing catastrophic forgetting enabling efficient adaptation through low-rank decomposition~\cite{hu2022lora, dettmers2023qlora}.

For API-based models (GPT's family), we employ provider-optimized fine-tuning protocols that leverage proprietary infrastructure for efficient parameter updates while maintaining model quality and inference speed.

\section{Documenting Cognitive Specialization in Current LLMs}
\label{sec:cognitive_specialization}


In this section, we empirically investigate the extent to which current LLMs exhibit \textbf{cognitive specialization}, a phenomenon where fine-tuning on forward code transformations enhances forward proficiency but erodes reverse reasoning capabilities.

\subsection{Experiment 1: Forward Transformation and Iterative Correction Assessment}
\subsubsection{Goal.}
Establish baseline performance of fine-tuned LLMs on obfuscation tasks and evaluate whether iterative auto-correction enables models to achieve performance comparable to specialized obfuscation tools.

\subsubsection{Experimental Setup.}
Each LLM presented in the previous section undergoes a task-specific fine-tuning. 
Each model is fine-tuned separately on variable renaming, dead code insertion, and string encryption datasets.

Concretely, for each model we obtain three fine-tuned versions, one for each transformation type.
Then, we used the evaluation dataset of 300 samples to query each fine-tuned LLM and ask them to generate obfuscated code.

The prompt instruction for the fine-tuning and obfuscated code generation is:

\begin{tcolorbox}[colback=gray!5, colframe=gray!25, coltitle=black, title=Prompt, boxrule=0.4pt, left=2pt, right=2pt, top=2pt, bottom=2pt]
\small\texttt{Obfuscate the following Java code by [Obfuscation Technique] while preserving its functionality.}
\end{tcolorbox}
where \textit{obfuscated technique} is either variable renaming, dead code insertion or literals encryption.

We evaluate these generated obfuscated code through two assessment: 
\bcircle{1} \textbf{semantic evaluation} using test case execution (pass rate, compilation success, runtime errors), and 
\bcircle{2} \textbf{syntactic evaluation} using CodeBLEU similarity computation. 
For syntactic analysis, we compute CodeBLEU scores $S \in [0,1]$ across four comparisons:

\begin{adjustbox}{width=.9\columnwidth,center}
\footnotesize
\noindent
\begin{minipage}[t]{0.48\linewidth}
\begin{align}
S_1 &= \text{CodeBLEU}(C_{\text{ft}}, C_{\text{orig}}) \label{eq:ft-orig} \\
S_2 &= \text{CodeBLEU}(C_{\text{ft}}, C_{\text{base}}) \label{eq:ft-base}
\end{align}
\end{minipage}
\hfill
\begin{minipage}[t]{0.48\linewidth}
\begin{align}
S_3 &= \text{CodeBLEU}(C_{\text{base}}, C_{\text{orig}}) \label{eq:base-orig} \\
S_4 &= \text{CodeBLEU}(C_{\text{ft}}, C_{\text{tool}}) \label{eq:ft-tool}
\end{align}
\end{minipage}
\end{adjustbox}

where $C_{ft}$ is fine-tuned model output, $C_{orig}$ is original code, $C_{base}$ is non-fine-tuned model output, and $C_{tool}$ is tool-generated reference obfuscation. Higher scores indicate greater syntactic similarity, with $S_4$ measuring alignment with ground-truth transformations.

Following the initial evaluation, we implement an iterative auto-correction loop (maximum 5 iterations) for failed attempts.
When obfuscated code fails to compile, to execute, or the test cases do not pass, error messages are fed back to the model with correction prompts to assess whether models can achieve tool-level performance through self-refinement (prompts are available in the supplementary materials).

\begin{figure*}[!ht]
    \centering
    \begin{minipage}[t]{0.47\textwidth}
        \centering
        \includegraphics[width=\linewidth, height=4cm, keepaspectratio]{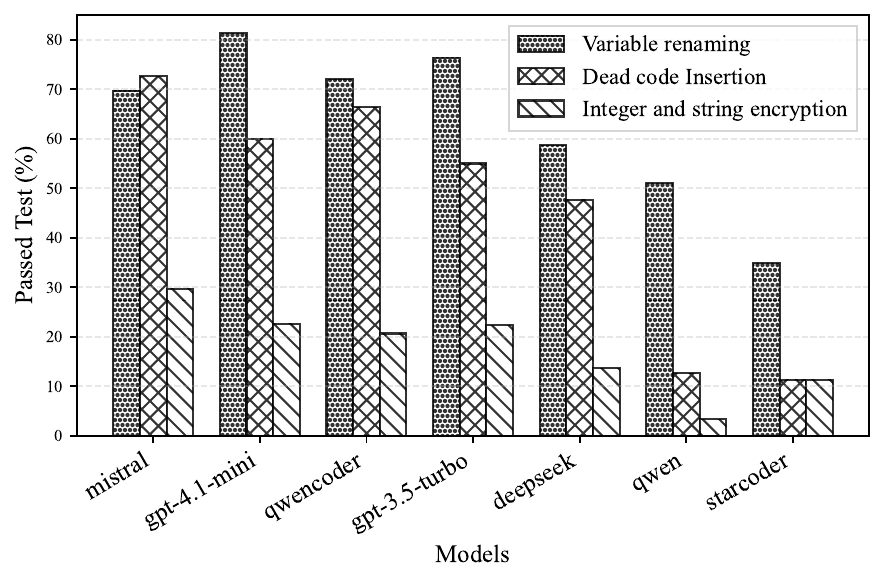}
        \caption{Fine-tuned model performance across code obfuscation techniques.}
        \label{fig:model_performance}
    \end{minipage}
    \hfill
    \begin{minipage}[t]{0.47\textwidth}
        \centering
        \includegraphics[width=\linewidth, height=5cm, keepaspectratio]{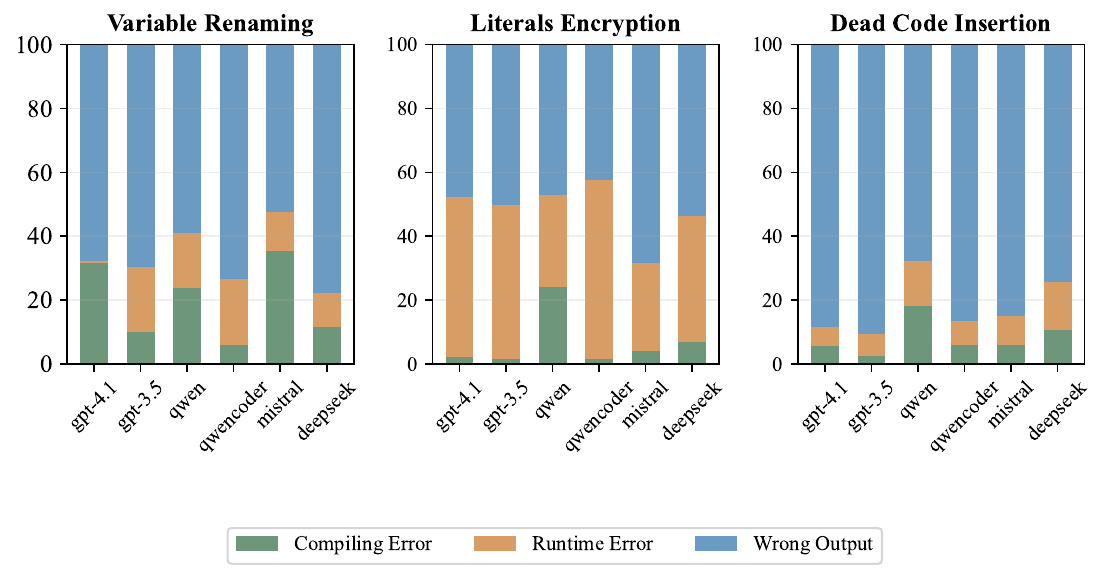}
        \caption{Error distribution of model-generated obfuscated code across three obfuscation tasks. Each bar shows the percentage of outputs that failed due to compilation error, runtime errors, or produced semantically wrong outputs.}
        \label{fig:errors_xp1}
    \end{minipage}
    \vspace{-0.5cm}
\end{figure*}

\noindent
\subsubsection{Results.}
\textit{Semantic Performance Hierarchy.} Our evaluation reveals systematic learning hierarchy across seven models and three transformation types: variable renaming (mean: 66.1\%, best: GPT-4.1-Mini at 81.3\%), dead code insertion (mean: 46.5\%, best: Mistral at 72.7\%), and string encryption (mean: 17.5\%, best: Mistral at 29.7\%) as illustrate in Figure~\ref{fig:model_performance}. The strong negative correlation between transformation complexity and model success (Spearman's $\rho = -0.91$, $p < 0.001$) indicates that models excel at simple pattern substitution but fail when transformations require algorithmic reasoning. Even top-performing models show 18--70 percentage point gaps from the deterministic tool baseline (100\%), confirming fundamental understanding limitations.

\textit{Semantic vs. Syntactic Competence.} Error (accross all models) analysis reveals that models can generate syntactically correct code but fail to preserve logical meaning. When models produce incorrect transformations, 69\% are ``wrong outputs''—code that compiles and runs but produces incorrect results. Only 4\% fail to compile, 27\% crash during execution as illustrated in Figure ~\ref{fig:errors_xp1}. This shows models learn code syntax well but lack semantic understanding. The problem worsens with complexity: wrong outputs increase from 62\% (for variable renaming) to 76\% (for string encryption). This is a hallmark of LLMs limit to truly understanding code semantic and code tranformation.

\textit{Pattern Learning Evidence.} CodeBLEU analysis confirms systematic pattern acquisition. Base models (Non-fine-tuned) produce outputs nearly identical to original code (Codebleu syntactic similarity 0.74--1.0) for complex transformations. Fine-tuning enables transformation behavior—outputs become dissimilar to original code (0.03--0.50) while achieving moderate similarity to tool references (0.49--0.78). StarCoder presents a deceptively perfect reproduction of the input (similarity = 1.0), so its results were excluded from the analysis. The observed negative correlation between transformation degree and semantic performance (Spearman’s $\rho = -0.73$, $p < 0.01$), along with the shift in similarity from original to obfuscated code, suggests that the model has acquired systematic but incomplete patterns during training.

\textit{Self-correction Limitations.} After five iterations of self-correction, improvements vary significantly across models and transformation types. For variable renaming, GPT-3.5-Turbo improves from 76\% to 96\%, and GPT-4.1-Mini from 81\% to 95\%, while Mistral shows only a modest gain from 70\% to 74\%. In contrast, string encryption remains consistently difficult: GPT-3.5-Turbo improves from 22\% to 47\%, GPT-4.1-Mini from 23\% to 48\%, and Mistral from 30\% to 34\%. Other models did not show statistically significant improvements. Across all models, performance gains plateau after the fifth iteration. A strong negative correlation is observed between self-correction effectiveness and transformation complexity (Spearman's $\rho = -0.85$, $p < 0.01$) indicates that LLMs favor syntactic correctness over semantic correctness.

\textit{Implications.}
While iterative feedback helps reduce compilation and execution errors, it often leads to an increase in incorrect outputs, indicating that only surface-level issues are being resolved. Results show that standard fine-tuning struggles with deeper algorithmic challenges. A consistent 69\% semantic failure rate, combined with limited self-correction on complex tasks, highlights the inadequacy of current approaches for robust code transformation. These findings underscore the need to explore alternative training methodologies better suited for handling semantic and structural complexity.



\subsection{Experiment 2: Generalization vs. Memorization Analysis}

\subsubsection{Goal.}
Assess whether models develop creative understanding of transformation principles or merely replicate training distribution patterns.

\subsubsection{Experimental Setup.}
We fine-tuned GPT-3.5-Mini and GPT-4.1-Mini---the best-performing models so far---on three  variable renaming strategies with controlled training equal distributions: 
\wcircle{1} sequential patterns (``i'', ``l'' repetitions), 
\wcircle{2} systematic naming (var1, var2, etc.), and 
\wcircle{3} custom schemes "\_[A-Z] random". 
Each model generated five outputs per 300 test samples (N=\num{1500} total generations per model) to assess pattern frequency replication versus novel strategy generation.

\subsubsection{Results.}
Fine-tuned models exhibit systematic replication of training distribution patterns rather than creative strategy generation.

\textit{Statistical Pattern Replication Dominates.} 
Chi-square tests confirm both models significantly deviate from uniform pattern distribution (p $<$ 0.001), directly mirroring training frequencies. GPT-3.5 follows training hierarchy: sequential patterns (42\%), systematic naming (40\%), custom schemes (16\%). GPT-4.1-Mini shows similar adherence but with different preferences: custom schemes (47\%), systematic naming (33\%), sequential patterns (19\%). Both models generate $<$2\% mixed patterns, indicating rigid category boundaries rather than flexible pattern interpolation.

When models generate novel patterns, they tend to be simple recombinations of elements seen during training rather than strategies driven by semantic intent. This reflects pattern replication rather than a principled understanding of obfuscation goals.

\begin{formal}
Pattern innovation analysis provides statistical evidence that LLMs operate as pattern replicators rather than creative reasoners. 
The low generation  (1.2\%) of mixed pattern from  fine-tuned, combined with mechanistic training frequency replication, demonstrates that fine-tuning produces statistical memorization instead of principled understanding of transformation objectives.
\end{formal}

\subsection{Experiment 3: Assessing Reversibility}

\subsubsection{Goal.}
Evaluate the core hypothesis of bidirectional reasoning by testing whether models fine-tuned with the obfuscation task
can perform deobfuscation.

\subsubsection{Experimental Setup.} 
All fine-tuned models were prompted to deobfuscate tool-generated obfuscated code using four approaches: 
\wcircle{1} simple prompting, 
\wcircle{2} few-shot learning, 
\wcircle{3} chain-of-thought reasoning, and 
\wcircle{4} augmented reasoning (reasoning + few-shot). 
Our evaluation employed three metrics:
\begin{itemize}
   \item \textbf{Syntactic Similarity}: $S(A,B) \in [0,1]$ measures CodeBLEU similarity between codes A and B (0=completely different, 1=identical)
   \item \textbf{Readability Score}: $R(C) \in [0,1]$ measures code readability quality (0=unreadable, 1=very readable)



    \item \textbf{Success Criterion}: Effective deobfuscation requires reducing syntactic similarity of the obfuscated code ($S(C_{\text{deobf}}, C_{\text{obf}}) \rightarrow 0$) while restoring readability to get close to the original code ($R(C_{\text{deobf}}) \rightarrow R(C_{\text{orig}})$).
\end{itemize}

\begin{figure*}[!ht]
\vspace{-0.4cm}
        \centering
        \includegraphics[width=\linewidth, height=5cm, keepaspectratio]{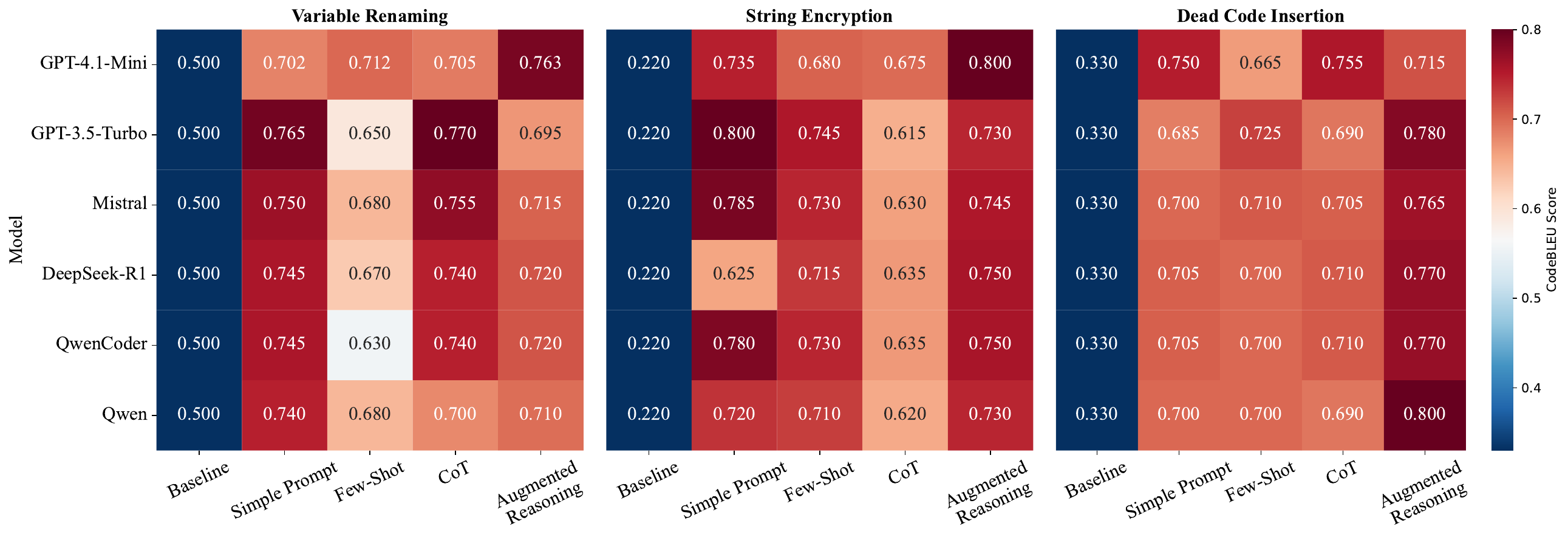} 
        \vspace{-0.4cm}
    \caption{Reversibility Evaluation. Baseline represent the Score between Original code and Tool obfuscated code}
    \label{fig:reversibility_testing}
    \vspace{-0.4cm}
\end{figure*}

\subsubsection{Results.}
\textit{Complete Deobfuscation Failure Across All Transformations.} 
Models demonstrate systematic inability to reverse learned transformations. Variable renaming shows $S(C_{deobf}, C_{obf}) \in [0.65-0.79]$, indicating models produce outputs nearly identical to obfuscated input rather than recovering original structure. String encryption maintains high similarity $S(C_{deobf}, C_{obf}) \in [0.61-0.8]$, failing to approach the desired similarity to original code $S(C_{obf}, C_{orig}) \approx 0.22$. Dead code insertion exhibits similar syntactic failure with $S(C_{deobf}, C_{obf}) \in [0.66-0.78]$, remaining far from baseline $S(C_{obf}, C_{orig}) \approx 0.33$ as illustrated in Figure ~\ref{fig:reversibility_testing}.

\textit{Dead Code Deobfuscation Reveals Complete Cognitive Disconnect.} 
Dead code results expose severe cognitive failure through a critical metric disconnect. Readability scores remain high for both obfuscated $R(C_{obf}) \in [0.78-0.88]$ and deobfuscated code $R(C_{deobf}) \in [0.78-0.88]$ because injected statements are syntactically valid. However, CodeBLEU similarity $S(C_{deobf}, C_{obf}) \in [0.66-0.78]$ expecting low, manual verification associate to number of token comparison of $C_{deobf}$ and $C_{obf}$ reveal models make virtually no modifications---they fail to remove any dead code despite explicit deobfuscation instructions. This demonstrates complete inability to distinguish functional from non-functional code elements.

\textit{Readability Analysis Confirms Systematic Failure.} 
Variable renaming attempts achieve $R(C_{deobf}) \in [0.55-0.69]$ compared to original $R(C_{orig}) = 0.79$, failing to restore meaningful identifiers from obfuscated baselines $R(C_{obf}) \in [0.55-0.65]$. String encryption deobfuscation produces $R(C_{deobf}) \in [0.19-0.32]$, showing no improvement over degraded obfuscated baselines $R(C_{obf}) \in [0.19-0.22]$. The consistent pattern---maintenance of obfuscated characteristics rather than recovery of original structure---confirms models replicate input patterns rather than understanding transformation semantics.

\textit{Prompt-Insensitive Results: $\Delta R \approx 0$.} 
Advanced prompting techniques fail to overcome fundamental limitations. Readability improvements remain negligible: $\Delta R = R(C_{advanced}) - R(C_{simple}) \in [0.01-0.05]$ across all strategies. Chain-of-thought reasoning with explicit instructions produces no meaningful changes, maintaining $S(C_{CoT}, C_{obf}) \approx S(C_{simple}, C_{obf})$, as illustrated in Figure ~\ref{fig:reversibility_testing}. This consistency confirms the limitation transcends methodological approaches, representing fundamental architectural constraints.

\textit{Cognitive Specialization: Forward Proficiency, Reverse Incapacity.} 
Models exhibit complete asymmetric capability profiles. Forward transformation success $P(T) > 0.85$ contrasts with reverse transformation success $P(T^{-1}) \approx 0$, where success requires at least $S(C_{output}, C_{target}) < 0.4$ and appropriate readability recovery. This asymmetry reveals that fine-tuning creates unidirectional neural pathways $T: C_{orig} \rightarrow C_{obf}$ while eliminating bidirectional reasoning capabilities $T^{-1}: C_{obf} \rightarrow C_{orig}$.

In addition to automated metrics, we manually verified a subset of outputs to ensure effective deobfuscation. We also compared token counts before and after transformation: obfuscation typically increases token count, while successful deobfuscation should reduce it, approaching the original size. These results are consistent across all evaluated models under a each prompting strategy, reinforcing the robustness of our evaluation framework.

\begin{formal}
Reversibility assessment provides definitive evidence of complete cognitive specialization. The stark asymmetry $P(T^{-1}) \approx 0$ despite $P(T) > 0.85$, combined with high similarity retention $S(C_{deobf}, C_{obf}) \in [0.61-0.79]$ and negligible prompting improvements $\Delta R \approx 0.01-0.05$, demonstrates that current fine-tuning approaches create irreversible unidirectional optimization.
\end{formal}

\section{Contrastive Fine-Tuning Solution}

\subsubsection{Goal.}The documented cognitive specialization in the previous section motivates the usage of Contrastive Fine-Tuning, a novel training methodology designed to enable bidirectional reasoning capabilities through semantically-aware multi-task learning.


\subsubsection{Experimental Setup.}
The systematic failure of fine-tuned models to perform inverse transformations ($P(T^{-1}) \approx 0$ despite $P(T) > 0.85$) demonstrates that current training approaches create unidirectional optimization pathways, limiting AI systems to pattern replication rather than semantic comprehension. CFT investigates whether multi-task learning can enable bidirectional reasoning capabilities while maintaining forward task performance, establishing the first methodology for overcoming cognitive specialization in fine-tuned language models.






\textit{CFT Theoretical Framework.} 
CFT aims to prevent cognitive specialization by training models on three complementary tasks that require semantic understanding. The approach combines three loss functions:
\vspace{-0.15cm}
\begin{equation}
\mathcal{L}_{CFT} = \mathcal{L}_{pos} + \mathcal{L}_{neg} + \mathcal{L}_{gen}
\vspace{-0.4cm}
\end{equation}

where:
\begin{itemize}
  \item $\mathcal{L}_{pos}$: Learns to recognize when original and obfuscated code are functionally equivalent
  \item $\mathcal{L}_{neg}$: Learns to detect when code has different functionality  
  \item $\mathcal{L}_{gen}$: Learns to generate obfuscated code 
\end{itemize}

This multi-task training encourages models to understand semantic relationships rather than memorizing surface patterns, enabling bidirectional reasoning capabilities.

\textit{Training Data Construction.} 
CFT uses balanced triplet datasets across transformation types. For open-source models, we construct \num{30000} instances (\num{10000} each for positive classification, negative classification, and obfuscation task generation). For API-based models, we use \num{3000} instances (\num{1000} per task). Positive classification involves semantically equivalent code pairs requiring binary equivalence decisions. Negative classification uses functionally different code to define semantic boundaries. Generation tasks apply forward transformations while preserving functionality.

We fine-tuned and evaluate CFT across six precedent representative models through three obfuscation tecniques.

\textit{Comparative Evaluation Framework.} 
CFT effectiveness is assessed through comparison against Standard Fine-Tuning (SFT) using and Bidirectional Fine-Tuning (BFT) using forward generation plus reverse deobfuscation tasks. Evaluation metrics include: forward obfuscation performance (semantic correctness, CodeBLEU similarity, readability preservation), reverse deobfuscation capability, and bidirectional consistency measured through transformation cycle analysis where success requires $S(T^{-1}) \approx S(T)$.


\begin{figure}[!ht]
        \centering
        \begin{adjustbox}{width=.8\columnwidth,center}
        \includegraphics[width=\linewidth, height=6cm, keepaspectratio]{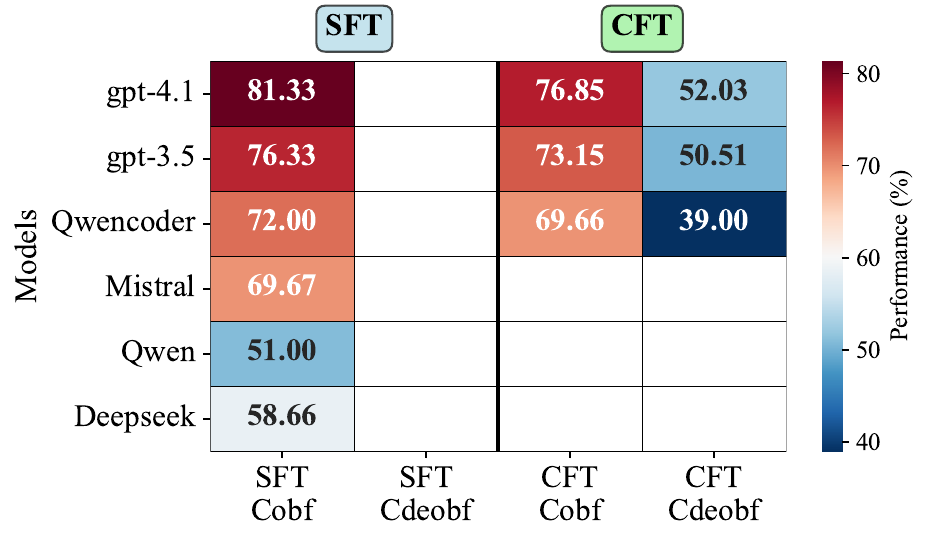} 
        \end{adjustbox}
        \vspace{-0.7cm}
    \caption{Comparison of single-task fine-tuning and Contrastive Fine-Tuning on forward and reverse obfuscation tasks. Empty cells mean that the model failed to reverse, reproducing the input in output.}
    \label{fig:bidirectional}
\end{figure}

\subsubsection{Results.}
CFT evaluation employs identical Experiment 4 protocols to enable direct comparison with cognitive specialization baselines.

\textit{Breakthrough Results: Variable Renaming Bidirectional Success.} 
CFT achieves categorical breakthrough in variable renaming transformations. GPT-4.1-Mini demonstrates \textbf{52.03\%} semantic success compared to 0\% SFT baseline, GPT-3.5-Turbo achieves \textbf{50.51\%} reverse capability and Qwencoder achieves \textbf{39\%} reverse capability, see Figure ~\ref{fig:bidirectional}. 

Manual verification confirms CFT generates meaningful identifiers (e.g., \texttt{currentCount}, \texttt{sumDigits}, \texttt{maxReplacements}, \texttt{totalNums}, \texttt{characters}) rather than obfuscated sequences (``I'', ``L'' patterns), demonstrating potential semantic understanding of variable purpose and context.

\textit{Architectural Capacity Hierarchy.} 
Commercial models demonstrate superior bidirectional emergence (GPT models: 50-52\%, QwenCoder: 39\%) while other open-source models show minimal improvement, revealing fundamental capacity constraints governing bidirectional reasoning emergence.

\textit{Forward Performance Preservation.}
CFT maintains forward transformation capability with $S(C_{forward}, C_{orig}) \in [0.42-0.51]$ versus SFT's $[0.42-0.50]$, confirming CFT overcomes unidirectional optimization pathways while preserving generation quality.

\textit{Transformation Complexity Hierarchy.} 
Variable renaming success while dead code insertion and literals encryption fail confirms semantic complexity constraints. Variable renaming requires minimal semantic understanding compared to complex dead code recognition or encryption logic, revealing current limitations in contrastive learning formulations.

\textit{Statistical Validation.}
Chi-square tests confirm significant SFT-CFT differences ($p < 0.001$), with effect sizes Cohen's $d > 3.0$ indicating extremely large practical significance. The categorical improvement (0\% $\rightarrow$ 50\%+) represents qualitative capability emergence rather than incremental enhancement.
\begin{formal}
CFT demonstrates the first successful emergence of bidirectional reasoning capabilities in fine-tuned language models, achieving 50-52\% reverse performance in variable renaming tasks compared to universal 0\% SFT failure. This breakthrough establishes proof-of-concept for overcoming cognitive specialization while revealing architectural and complexity constraints governing bidirectional reasoning emergence.
\end{formal}

\subsection{Computing Infrastructure}
Open-source model experiments were conducted on NVIDIA A100-SXM4-40GB clusters: 4 GPUs for fine-tuning and 3 GPUs for inference, using CUDA 12.8 and driver version 570.133.20. API-based models (GPT-3.5-Turbo, GPT-4.1-Mini) were fine-tuned and evaluated through provider APIs with a total computational budget of approximately €800.
\vspace{-0.3cm}

\section{Related Work}
\label{related_work}

\subsection{Fine-Tuning Limitations and Specialization}
~\cite{kirkpatrick2017overcoming} documented catastrophic forgetting in neural networks, where task-specific training eliminates previously acquired capabilities. ~\cite{french1999catastrophic} identified similar patterns in connectionist networks, establishing that sequential learning creates interference between tasks. 

Recent parameter-efficient approaches like ~\cite{hu2022lora}    attempt to mitigate catastrophic forgetting through selective parameter updates. However, these focus on preserving forward task performance rather than enabling bidirectional reasoning capabilities. 

Our findings extend prior work by showing that fine-tuning not only preserves parameters but also enforces unidirectional optimization, which hinders bidirectional reasoning and limits understanding.

\textit{Limitations of Existing Evaluation and Motivation.} 
Prior work such as \cite{chen2021evaluating} and \cite{austin2021program} established benchmarks and demonstrated strong generation performance, but focused solely on forward tasks—potentially conflating pattern matching with genuine understanding. Our bidirectional framework introduces more rigorous criteria to distinguish true code comprehension from surface-level replication.

\textit{Contrastive Learning for Code Understanding.} 
Contrastive learning has proven effective for semantic representation learning, particularly in vision~\cite{chen2020simple, he2020momentum}. While existing methods target representation quality, we introduce a novel contrastive fine-tuning (CFT) framework tailored to code transformation tasks. By leveraging semantic triplets that encode transformation relationships, CFT explicitly targets bidirectional reasoning—marking the first adaptation of contrastive learning for code understanding through transformation-based supervision.

\subsection{Bidirectional Learning}
Cycle-consistency in vision tasks~\cite{zhu2017unpaired} shows that bidirectional constraints improve learning, but typically require explicit forward and backward training. In contrast, our method uses multi-task contrastive learning to encourage bidirectional reasoning—without needing reverse examples—marking a step forward in bidirectional learning for code.

\section{Conclusion}
\label{sec:conclusion}

This work introduces bidirectional reasoning as a meaningful benchmark for assessing genuine understanding in language models. We identify cognitive specialization as a systematic limitation induced by standard fine-tuning, and propose Contrastive Fine-Tuning (CFT) as a solution. CFT enables the emergence of bidirectional reasoning (39–52\% reverse performance) through multi-objective learning that jointly encourages semantic analysis and pattern recognition.

Although current results are limited to simple code transformations, CFT provides initial evidence that language models can move beyond pattern replication toward more robust semantic understanding. These findings lay the groundwork for future research into understanding-oriented AI systems, offering both a theoretical perspective and empirical support for this direction.

\bibliographystyle{ACM-Reference-Format}
\bibliography{references}

\end{document}